\pdfoutput=1

\documentclass[11pt]{article}

\usepackage{acl}

\usepackage{times}
\usepackage{latexsym}
\usepackage{multirow}

\usepackage{pdfpages}

\usepackage[T1]{fontenc}

\usepackage[utf8]{inputenc}

\usepackage{microtype}

\title{Context Enhanced Short Text Matching using Clickthrough Data}

\def\correspondingauthor{\thanks{$^*$Corresponding authors(\href{mailto: yang.yujiu@sz.tsinghua.edu.cn}{yang.yujiu@sz.tsinghua.edu.cn}, \href{mailto: haiyunjiang@tencent.com}{haiyunjiang@tencent.com}).}}

\author{Mao Yan Chen$^{1,2}$, Haiyun Jiang$^{2}$\correspondingauthor  \and Yujiu Yang$^{2*}$ \\
        $^1$Tsinghua Shenzhen International Graduate School, Tsinghua University \\ $^2$Tencent AI Lab \\
        \texttt{chenmaoy19@mails.tsinghua.edu.cn} \\
        \texttt{haiyunjiang@tencent.com} \\
        \texttt{yang.yujiu@sz.tsinghua.edu.cn}}

\begin{document}
\maketitle

\begin{abstract}
The short text matching task employs a model to determine whether two short texts have the same semantic meaning or intent. Existing short text matching models usually rely on the content of short texts which are lack information or missing some key clues. Therefore, the short texts need external knowledge to complete their semantic meaning. To address this issue, we propose a new short text matching framework for introducing external knowledge to enhance the short text contextual representation. In detail, we apply a self-attention mechanism to enrich short text representation with external contexts. Experiments on two Chinese datasets and one English dataset demonstrate that our framework outperforms the state-of-the-art short text matching models.
\end{abstract}

\section{Introduction}
Short text matching is an essential task that has been applied in question answering~\cite{berger2000bridging}, paraphrase identification~\cite{socher2011dynamic} and information retrieval~\cite{huang2013learning}. In recent years, deep neural networks achieve surprising performance in this field. 
We can roughly classify deep text matching models into two types: 1) representation-based text matching~\cite{huang2013learning,shen2014learning} and 2) interaction-based text matching~\cite{wang2015learning,wang2017bilateral,devlin2018bert}. 
The interactive-based framework is usually performs better than the representation-based framework.
Though interactive-based framework achieved very promising results, their performance still suffers from the lack of enough contextual information since words or expressions in a short text usually have ambiguous meanings.

Especially in Chinese scenarios, both character-level and word-level tokenization introduce serious semantic information errors or missing.
Recent studies show that encoding multi-granularity information~\cite{lai2019lattice,chen2020neural} and word sense information~\cite{lyu2021let} into sentences can mitigate this problem.
They further improve the performance, while this word-level information reinforces is still helpless in many cases that need relevant sentence-level contextual information supplement.

\begin{figure}[tbp]
    \centering
    \includegraphics[width=\linewidth]{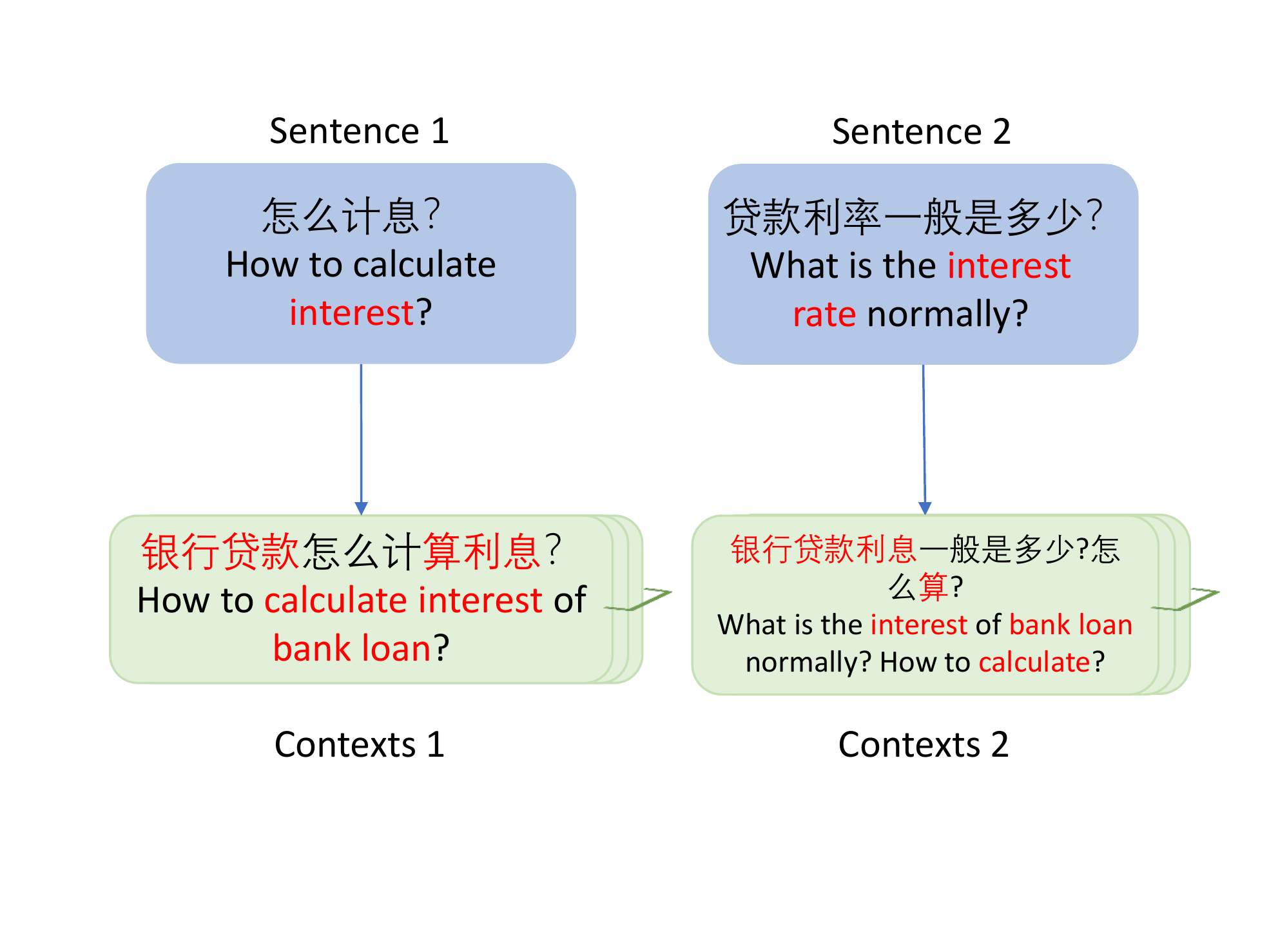}
    \caption{Sentence 1 and 2 are the short texts. Context 1 and 2 are their crawled contexts from search engine. Highlighted spans are the matched part between short texts and contexts.}
    \label{fig:sent_info}
\end{figure}

As seen in Figure~\ref{fig:sent_info}, sentences 1 and 2 refer to the same question but the word-level semantic information is not enough to connect them. Therefore, we take the original sentences as queries to search related contexts by search engines. The retrieved contexts usually contain enough contextual information to relate the two original short texts. In this case, both short texts refer to "interest of bank loan", where the matching model could easily classify them to be matched.

From this insight, we propose a context-aware BERT matching model (CBM) for short text matching, which enrich the semantic representation of a short text by external semantic-related sentences, instead of word-levle knowledge. 
As seen in Figure \ref{fig:sent_info}, both sentences have multiple related contextual sentences\footnote{We denote a contextual sentence as a context for short.}. CBM selects the needed contexts and updates the short text representation according to the context-enhanced attention mechanism. Our experiments on two Chinese datasets and one English dataset show that our model achieves new state-of-the-art performance. Our contributions are three folds:
\begin{itemize}
    \item We are the first to propose a framework that enhances short text representation by external sentence-level knowledge.
    \item We crawled a huge amount of contextual sentences for the three commonly used benchmark datasets, which benefits future research.
    \item We design a simple but efficient model to utilize the sentences for short text representation reinforcement. Experiments show that our model achieves new SoTA performance.
\end{itemize}

\section{Framework}

Given two sentences, $S_a=\{s_a^1, s_a^2, ..., s_a^i, ..., s_a^n\}$ and $S_b=\{s_b^1,s_b^2, ..., s_b^j,...,s_b^m\}$, we aim to decide whether two sentences have the same semantic meaning. $s_a^i$ and $s_b^j$ denotes the $i$-th and $j$-th token in sentence $a$ and $b$, respectively. Different from existing methods, we not only use the sentences in datasets, but also utilize the external sentences crawled from search engines to enhance the context. Each sentence $S_i$ has a set of contexts: $C_i= \{c_i^1,c_i^2,..., c_i^j, ..., c_i^n\}$, where $c_i^j$ represents the $j$-th context for sentence $S_i$. 

Our framework has three modules: 1) Contexts Crawler, 2) Context Selector, and 3) Context-enhanced Text Matcher.

\subsection{Context Crawler}
For each sentence $S_i$, we obtain the set of contexts $C_{i'}$ corresponding to $S_i$ by crawling the search engine results. The retrieved contexts $C_{i'}$ are noisy and dirty, so we first remove the noise by pre-processing and perform a regular cleaning. Also, all contents related to personal information is removed. Finally, we will have a clean context set $C_i$ for each sentence $S_i$.

\subsection{Context Selector}
First, we use BERT baseline model to perform semantic similarity task for each pair of sentence and context, $S_a$ with $c_b^j$ or $S_b$ with $c_a^j$. Aftrer that, each pair of a sentence and a context has a similarity score of $d_i^j$, higher means higher semantic similarity, lower means lower semantic similarity. For instance, $d_a^j$ is the similarity score for the pair of $S_a$ and $c_b^j$. For all positive samples ($S_a$ and $S_b$ are semantically matched), we use the hyperparameter $d_a$ to classify the context and sentence pairs into similar or dissimilar. All $d_i^j>d_a$ is similar and others are dissimilar. Otherwise, for all negative samples ($S_a^j$ and $S_b^j$ are not semantically matched), we use the hyperparameter $d_b$ to classify the context and sentence pairs into similar and dissimilar, with all $d_i^j>d_b$ being similar and the rest being dissimilar.

For all positive samples, $S_a^{+}$ and $S_b^{+}$, we want the context of $S_a^{+}$ to have similar semantic information as $S_b^{+}$. Also, we want the context of $S_b^{+}$ to have similar semantic information as $S_a^{+}$. On the contrary, for all negative samples, $S_a^{-}$ and $S_b^{-}$, we expect the contexts of $S_a^{-}$ to be semantically dissimilar to $S_b^{-}$. It is also expected that the contexts of $S_b^{-}$ are not semantically similar to $S_a^{-}$.

However, we do not have ground truth labels for test set to make the context selection by labels. Therefore, we construct a context selector to determine whether we want to use the context based on the semantic information of the two sentences and the context.

When we construct contexts for the above positive and negative samples, using similar semantic contexts for positive samples and dissimilar semantic contexts for negative samples. We first put this data as pseudo labels into a BERT classifier for training. The inputs are  $S_a$, $S_b$, $c_i^j$, where $i\in[a,b]$, and the model output will be $[0,1]$, indicating whether this context will be used or not. Finally, the context selector integrates all relevant contexts into one context set, $\hat{C}$.

\subsection{Context-enhanced Text Matcher}
First, we encode $S_a$ and $S_b$ by sentence BERT to obtain their embedding, $h_a$ and $h_b$. Then, we use context BERT model to encode $\hat{c}_a$, $\hat{c}_b$ to obtain the embeddings of the contexts, $h_a^c$ and $h_b^c$, respectively.
Afterward, we concatenate $h_a$, $h_b$, $h_a^c$ and $h_b^c$ together and input them into a 3-layer Transformer model. Finally, we obtain the representation $h_a$, $h_b$, which are the final context-enhanced text representation of $S_a$ and $S_b$.

\begin{figure*} 
    \centering
    \includegraphics[width=\textwidth]{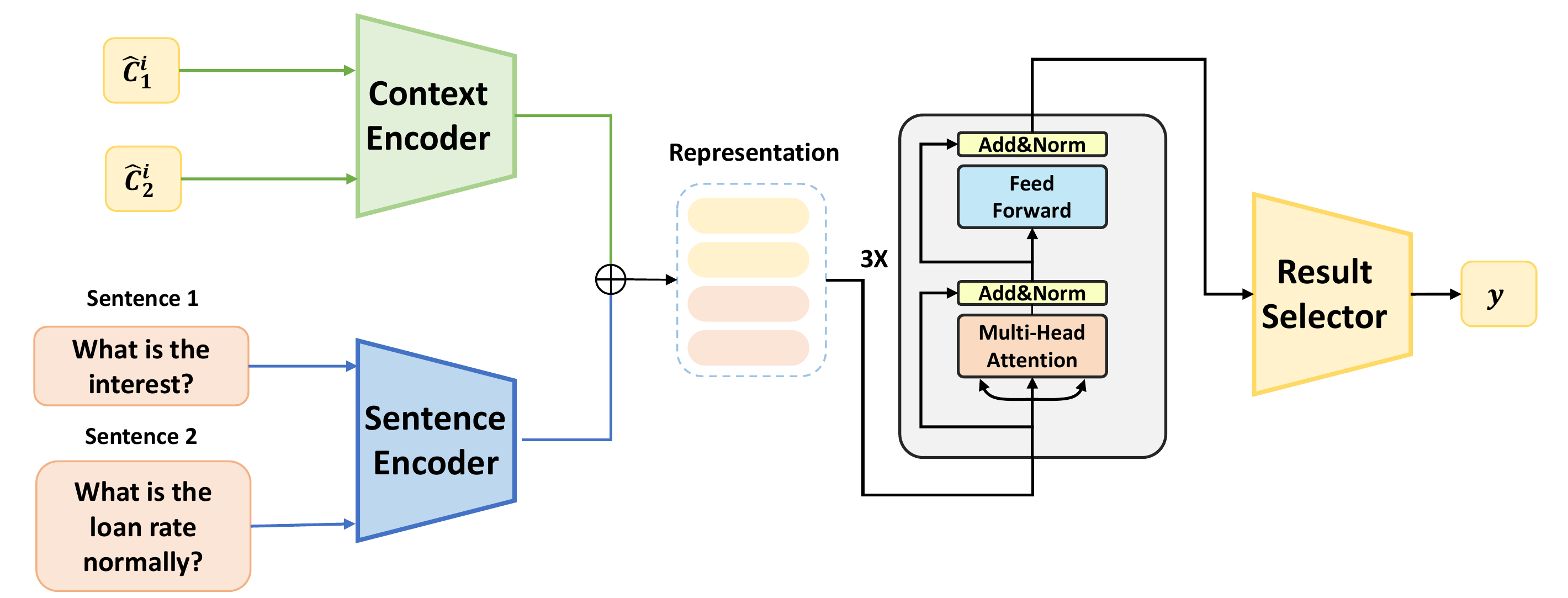}
    \caption{Framework}
    \label{fig:pipeline}
\end{figure*}

\subsection{Matching Classifier}
Our model predict predict the text similarity of two context-enhanced text representations.
\begin{equation}
h_{final}=[h_a;h_b;|h_a-h_b|] \label{1}
\end{equation}


\begin{equation}
p_i=FFN(h_{final}) \label{2}
\end{equation}
where FFN(·) is a feed forward network with two hidden layers and a output layer, a relu activation after each hidden layer.

For each training sample $\{S_a, S_b, y\}$, we aim to minimize the BCE loss:

\begin{equation}
\mathcal{L}=-\sum_{i=1}^{N}(ylog(p_i)+(1-y)log(1-p)) \label{3}
\end{equation}
where $y\in\{0, 1\}$ is the label of the i-th training sample and $p\in\{0, 1\}$ is the prediction of our model taking the sentence pair as input.

\subsection{Result Selector}
Since not every pair of short texts need context enhancement, for those pairs have high confidence with BERT baseline, we will keep the results and logits. We set the output logits of BERT baseline and our model to be $\hat{y}_i$ and $\bar{y}_i$, respectively. Then, the final result will be as follow:
\begin{equation}
y_i = \hat{y}_i + \bar{y}_i - 1
\end{equation}
where $y_i\in\{0, 1\}$ is the final predicted label of i-th sample, and $y_i$ equal to 1 if $y_i$ is larger than or equal to 0.5. Otherwise, $y_i$ will be set to 0.
\section{Experiments}

\begin{table}
\centering
\resizebox{\linewidth}{!}{%
\begin{tabular}{lllll}
\hline
\multirow{2}{*}{Model} & \multicolumn{2}{l}{BQ} & \multicolumn{2}{l}{LCQMC} \\ \cline{2-5} 
                       & ACC        & F1        & ACC         & F1          \\ \hline
BERT-Baseline*         & 84.8       & 84.6      & 87.6        & 88.0        \\
ERNIE 2.0              & 85.2       & -         & 87.9        & -           \\
LET-BERT               & 85.3       & 84.98     & 88.38       & 88.85       \\
ZEN 2.0 Base           & 85.42      & -         & 88.71       & -           \\
GMN-BERT               & 85.6       & 85.5      & 87.3        & 88.0        \\
Glyce+bERT             & 85.8       & 85.5      & 88.7        & 88.8        \\
ROBERTA-wwm-ext-large* & 85.79      & -         & \textbf{90.4}        & -           \\
Ours-BERT*             & \textbf{86.16}      & \textbf{87.44}     & \textbf{88.8}        & \textbf{89.1}        \\
Ours-RoBERTa*          & \textbf{86.66}      & \textbf{86.69}     & \textbf{89.2}           & \textbf{88.8}           \\ \hline
\end{tabular}%
}
\caption{Experiments on BQ and LCQMC Datasets. * marks the results reproduced by us. }
\label{tab:exp}
\end{table}

\subsection{Dataset}
We conduct our experiments on Bank Question (BQ) \cite{chen2018bq},  large-scale Chinese question matching corpus (LCQMC) \cite{liu2018lcqmc} and the Quora Question Paraphrasing corpus (QQP) datasets for semantic textual similarity task. BQ is a large-scale domain-specific Chinese corpus for sentence semantic matching. It is collected from customer service logs by a Chinese bank. LCQMC is a large-scale chinese question matching corpus. It focuses on intent matching rather than paraphrase.

\begin{table}[bp]
\resizebox{\linewidth}{!}{%
\begin{tabular}{|l|l|l|}
\hline
Model          & ACC    & F1     \\ \hline
Ours           & 86.16 & \textbf{87.44} \\ \hline
Ours+RoBERTa   & \textbf{86.66} (+0.5) & 86.69 (-0.75) \\ \hline
Ours-share     & 85.79 (-0.37) & 87.10 (-0.34)  \\ \hline
Ours-cs+random & 84.82 (-1.34) & 84.40 (-3.04)  \\ \hline
Ours-cs+topk   & 84.23 (-1.93) & 84.20 (-3.24)  \\ \hline
Ours-rs   & 85.45 (-0.71) & 85.37 (-2.07)  \\ \hline
\end{tabular}%
}
\caption{Ablation studies on BQ dataset}
\label{tab:ablation}
\end{table}

\subsection{Experiments}
\begin{itemize}
\item[$\bullet$]BERT-Baseline: A chinese pretrained BERT, called Chinese-BERT-wwm, provided by \cite{cui2019pre}.
\item[$\bullet$]ERNIE 2.0:  A continual pre-training framework named ERNIE 2.0 which incrementally builds pre-training tasks and then learn pre-trained models on these constructed tasks via continual multi-task learning. \cite{sun2021ernie}
\item[$\bullet$]LET-BERT\cite{lyu2021let}: A Linguistic knowledge Enhanced graph Transformer (LET) to deal with word ambiguity using HowNet.
\item[$\bullet$]ZEN 2.0 Base\cite{song2021zen}: An updated n-gram enhanced pre-trained encoder on Chinese and Arabic.
\item[$\bullet$]GMN-BERT: A neural graph matching method (GMN) for Chinese short text matching.
\item[$\bullet$]Glyce+BERT\cite{meng2019glyce}: Glyce provide glyph-vectors for logographic language representations. 
\item[$\bullet$]RoBERTa-wwm-ext-large: A chinese pretrained RoBERTa model which is also provided by \cite{cui2019pre}.
\end{itemize}
In Table \ref{tab:exp}, both Bert Baseline and our model are the results of our tuning of the hyperparameters to the best. All other experimental results are using the best results on the corresponding paper. In comparison, our model results outperform all baselines on the BQ dataset and outperform the previous best model by nearly 2\% in F1 values. On the LCQMC dataset, we also achieve the state of the art results.

\subsection{Details}
The BERT models used in our experiments are the BERT-wwm-ext version provided by HIT University. Learning rate, epoch size, sequence size and batch size and context size are 3e-5, 3, 512, 64 and 3, respectively. All experiments are run on V100 graphics cards.

\subsection{Ablation Studies}
\textbf{Ours+RoBERTa}: Replacing BERT model in our framework with RoBERTa model. \\
\textbf{Ours-share}: Removing share parameter mechanism between context encoder and sentence encoder. \\
\textbf{Ours-cs+random}: Removing context selector module and randomly choosing k contexts. \\
\textbf{Ours-cs+topk}: Removing context selector module and choosing top k relevant contexts. \\
\textbf{Ours-rs}: Removing result selector module. \\
As seen in Table \ref{tab:ablation}, removing the context selector will hurt the performance significantly because the unfiltered contexts contain serious noise. If we randomly select k contexts, then the model could not take advantage of the context information since the contexts may be irrelevant. However, when we select K most relevant contexts, due to the properities of the search engine, contexts will be exactly the same or similar to short texts. Therefore, the contexts will turn out to be useless for the model. As a result, the ablation studies proves that the context selector module can effectively filter the noisy contexts and provide high-quality contexts for each short text.\\
Our model shares the parameters in context encoder and short text encoder, then they could encode the contexts and short texts into the same semantic space. Finally, the ablation studies demostrate that sharing parameters boosts the performance and efficiency of the model.

\nocite{meng2019glyce,chen2020neural,lyu2021let,cui-etal-2020-revisiting,song2021zen,sun2021ernie,vaswani2017attention,devlin2018bert,liu2018lcqmc,chen2018bq}
\section{Conclusion}
\label{sec:bibtex}
In this work, we proposed a novel external knowledge enhanced BERT for Chinese short text matching. Our model takes two sentences and some related contexts as input and integrates the external information to moderate word ambiguity. The proposed method is evaluated on two Chinese benchmark datasets and obtains the best performance. Theablation studies also demonstrate that both semantic information and multi-granularity information are important for text matching modeling.

\bibliography{anthology,custom}
\bibliographystyle{acl_natbib}




\end{document}